%% file: root.tex
\documentclass[letterpaper, 10 pt, conference]{ieeeconf}  

\IEEEoverridecommandlockouts                              

\usepackage{graphicx} 
\usepackage{booktabs}

\usepackage[T1]{fontenc}
\usepackage{array}

\usepackage[export]{adjustbox}
\usepackage[colorlinks = true,
            linkcolor = black,
            urlcolor  = blue,
            citecolor = black,
            anchorcolor = black]{hyperref}
\usepackage{siunitx}
\usepackage{todonotes}
\usepackage{mathtools}
\usepackage{graphicx}
\usepackage{caption}

\usepackage{subcaption}
\usepackage{pgf}
\usepackage{layouts}
\usepackage{multirow}
\usepackage{url}
\newcommand{\etal}{~\emph{et al.}}

\input{python_results}

\usepackage{etoolbox}
\makeatletter
\patchcmd{\@makecaption}
  {\scshape}
  {}
  {}
  {}
\makeatletter
\patchcmd{\@makecaption}
  {\\}
  {.\ }
  {}
  {}
\makeatother

\newcommand{\algName}{CloudTrack}
\title{\LARGE \bf
\algName: Scalable UAV Tracking with Cloud Semantics
}

\author{Yannik Blei$^{1}$, Michael Krawez$^{1}$, Nisarga Nilavadi$^{1}$, Tanja Katharina Kaiser$^{1}$ and Wolfram Burgard$^{1}$
\thanks{$^{1}$All authors are with the Department of Computer Science \& Artificial Intelligence, University of Technology Nuremberg, Germany.}%
}

\begin{document}

\maketitle
\thispagestyle{empty}
\pagestyle{empty}

\begin{abstract}
Nowadays, unmanned aerial vehicles (UAVs) are commonly used in search and rescue scenarios to gather information in the search area. The automatic identification of the person searched for in aerial footage could increase the autonomy of such systems, reduce the search time, and thus increase the missed person's chances of survival. In this paper, we present a novel approach to perform semantically conditioned open vocabulary object tracking that is specifically designed to cope with the limitations of UAV hardware. Our approach has several advantages: It can run with verbal descriptions of the missing person, e.g., the color of the shirt, it does not require dedicated training to execute the mission, and can efficiently track a potentially moving person. Our experimental results demonstrate the versatility and efficacy of our approach. 
We publish the methods source code at \url{https://github.com/yblei/CloudTrack}.

\end{abstract}

\section{INTRODUCTION}

Visual detection and tracking of objects is an integral part of many modern-day UAV applications. Traditional object detection approaches~\cite{Redmon_2016_CVPR, ren2016faster, cai2019cascade} based on Convolutional Neural Networks (CNNs) require careful collection and curation of data to reach acceptable levels of performance. In search and rescue (SAR) operations, the object of interest is usually unknown when developing the system. Thus, traditional approaches are often inapplicable or limited to a small set of pre-trained classes. Instead, in many SAR scenarios, a semantic description is given, e.g., ``The lost person is male and wears a gray shirt'' or ``The person was last seen driving in a red car near a forest''. Even though modern Vision Language Models (VLMs) can exploit such information, they are typically inapplicable in UAVs due to a limitation of onboard hardware resources. Further, comparably low frame rates of such pipelines limit the ability to track objects of interest across multiple frames.

Other object trackers, such as DaSiamRPN~\cite{zhu2018Distractoraware}, deliver acceptable frame rates even on limited hardware, e.g., a Raspberry Pi 4. Such lightweight trackers are usually initialized by a bounding box around the target object and track it across upcoming frames. However, the object of interest itself must first be identified by other means. 
Our objective is an approach for semantically conditioned object detection and tracking, applicable in real-time on UAVs without further training or fine-tuning. 

\begin{figure}[t]
    \centering
    \includegraphics[width=\linewidth]{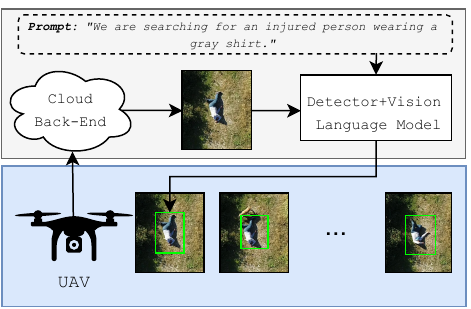}
    \caption{Detection of a person in UAV footage based on a semantic description. An image is recorded by the UAVs onboard camera. The image is then processed in the cloud by an open vocabulary object detector and a VLM. If the VLM confirms a match with the description, the object is tracked by an onboard real-time object tracker.}
    \vspace{-4mm}
\label{fig:enter-label}       
\end{figure}

We approach this goal by presenting \algName, an open vocabulary (OV) object detector and tracker. Our system consists of two parts: A tracking front-end delivers close to real-time performance on common UAV companion computers and has been evaluated on two platforms. An AI back-end leverages multiple foundation models to provide object detection based on higher-level semantics. This client-server architecture allows the potential application of multiple UAVs to reduce the required time for covering a given area. We further extend two publicly available benchmarks, VOT22~\cite{Kristan2022a} and SARD~\cite{sambolek_search_2021}, by semantic information and release the data as a part of this work. We test our approach on both datasets. Extensive experiments validate our approach.\footnote[2]{Video available at \url{https://youtu.be/GtfX8S_oMAE}}


\textbf{In summary, we make the following contributions:}

\begin{enumerate}
    \item We propose a novel approach for online OV object detection and tracking with elevated semantic accuracy, respecting the limitations of drone hardware.
    \item We release the code, including a ROS implementation.
    \item We release referring expressions for the VOT22 benchmark and semantic ground truth information for the SARD dataset.  
    \item We conduct an extensive evaluation using common UAV companion computers and demonstrate improvement over two state-of-the-art detection baselines.
\end{enumerate}

\begin{figure*}[t]
    \centering
    \includegraphics[width=\textwidth]{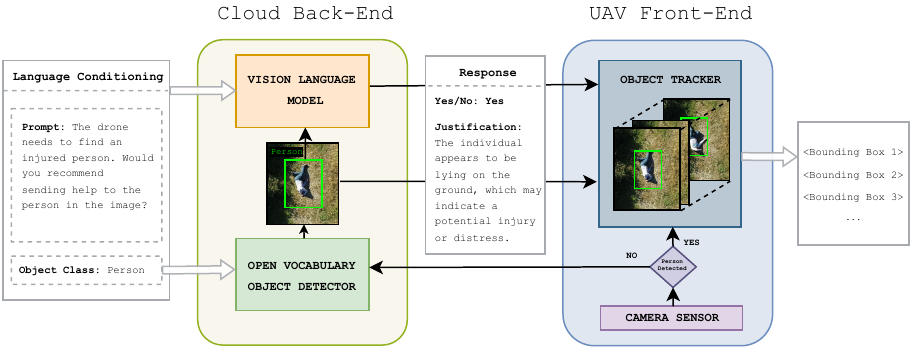}
    \caption{Overview of the object tracking pipeline. We run an OV object detector and a vision language model in the back-end. UAVs are equipped with a camera and a lightweight object tracker. The onboard object tracker is initialized via the back-end. It ensures close to real-time frame rates on limited hardware. Multiple agents can share the same back-end to achieve better scalability in search and rescue scenarios.}
    \label{figure:flowchart}
    \vspace*{-4mm} 
\end{figure*}

\section{RELATED WORK}
\subsection{Detection and Tracking for UAVs}
Numerous approaches for person detection and tracking in SAR missions have been proposed. The focus often lies on enabling onboard processing, for instance, by deploying lightweight models~\cite{rizk2021toward, jung2022improved, zaman2023human} like YOLO~\cite{Redmon_2016_CVPR}. Xu\etal~\cite{xu2023collaborative} tackle the problem of limited edge computing with a collaborative learning scheme. A lightweight onboard network sends noisy person detections to the mission center for verification. Using the corresponding feedback, the UAV network is improved over time. 

A similar idea is used by Zhou and Liu~\cite{zhou2021long} to improve an onboard person tracker through human operator feedback provided by a gaze tracking system. Our method also relies on a connection to a ground station for tracking initialization but we deploy a VLM instead of a human operator. Generally, detection and tracking can be both performed on an edge device. For instance, Lo\etal~\cite{lo2021dynamic} deploy YOLO for detection and a Kalman filter for general object tracking. Similarly, Chen\etal~\cite{chen2023fully} detect pedestrians with YOLO and propose a motion model-based tracking algorithm. Both methods deliver acceptable performance on closed, pre-trained object sets. However, current drone computing is too restrictive for full open-world detection.

\subsection{Model-free Object Tracking}
Model-free approaches aim at tracking arbitrary objects in an image stream given that the target object is marked in the first frame. Thus, they require an appropriate detection method for initialization.
MIL-Track~\cite{babenko2009visual} implements the tracking-by-detection strategy. It online learns a discriminative appearance model of the tracked object from positive and negative image patches.  The key insight there is to consider multiple overlapping positive patches as one sample. Similarly, CSRT~\cite{lukezic2017discriminative} is a discriminative tracking approach that uses spatial reliability maps which allow considering larger background regions around the object and improve tracking of objects with irregular shapes. 

GOTURN~\cite{held2016learning} deploys a simple CNN to find the tracking target in the current image given the detection in the previous frame. Zhu\etal~\cite{Zhu_2018_ECCV} propose the siamese network-based DaSiamRPN-tracker and a distractor-aware training strategy that improves target tracking among visually similar objects. Yan\etal~\cite{yan2021lighttrack} use meta-parameter search to find efficient network architectures for object tracking.

\subsection{Open-Vocabulary Detection and Tracking}
Open vocabulary or open-world methods can detect, segment, or track a wide domain of object classes according to a textual user description. This is made possible by pre-training large vision and text encoders on internet-scale datasets of text-image pairs, e.g., CLIP~\cite{radford2021learning}, and reusing the components for detection or segmentation.  Dedicated models for detection like OWL-ViT~\cite{minderer2022simple} and Grounding DINO~\cite{liu2023grounding} have been proposed and also recent generalist VLMs like PaliGemma~\cite{beyer2024paligemma} or LLaVA~\cite{liu2024improved} offer OV object detection.

The object-tracking community also harnessed the capabilities of OV models.
OVTrack~\cite{li2023ovtrack} deploys a Faster \linebreak R-CNN~\cite{ren2016faster} backbone and replaces its classification head with an image, a text, and a tracking head. During training, the text and image heads are aligned with the corresponding CLIP embeddings. Chu\etal~\cite{chu2024zero} use pre-trained VLMs to detect and segment referred objects. Closely related is the problem of semantic video segmentation, where recent approaches~\cite{cheng2023tracking, heigold2023video, wu2023GLEE}  have also made the jump to the OV domain.

\section{METHODS}


\subsection{Overview} 
The overall objective of this work is to detect and track objects of a class, defined by a verbal description, using one or potentially multiple UAVs. We call instances of this class objects of interest. The system must work in an OV manner since the object of interest in most use cases is typically unknown during development time. Even though the system supports a wide range of potential classes, we assess its capabilities in the scope of a SAR mission. Potential prompts in this setting are {\tt "An injured male person wearing a gray shirt."} or {\tt "A red SUV, parked in the forest."}. When an instance is found, the UAV tracks the object and executes a predefined behavior. For example, in a SAR mission, it could follow the object awaiting feedback from a human supervisor.

The proposed approach consists of two main modules as depicted in \autoref{figure:flowchart}. The back-end (Section \ref{sec:method_backend}) runs on a server and performs OV object detection to initialize or re-initialize the front-end tracker (Section \ref{sec:method_frontend}) running on the UAV. We assume that a network connection between the UAV and the server exists.

\subsection{Object-Detection Back-End}
\label{sec:method_backend}
The back-end consists of an OV object detector and a VLM. 
Notably, the OV object detector evaluated in this work demonstrates strong performance even when identifying small instances relative to the overall image size. It struggles, however, to understand the deeper semantics of a prompt as it is common in SAR missions. Thus, prompting the system with, e.g., {\tt "An injured male person wearing a gray shirt."}, results in bounding boxes of all individuals in the image, independent of potential signs of injury, a match in cloth color or gender. This leads to a high number of false positives and could overwhelm human supervisors. VLMs on the other hand feature strong capabilities of semantic understanding. On the contrary, they struggle to correctly identify small objects in the image. This leads to unexpected behavior, such as random answers and hallucinations. 

Therefore, we prompt the OV object detector with a simpler superset of the class of interest (superset class). For example, when searching for {\tt "an injured person"}, this could be {\tt "person"} or {\tt "human"}.
We then crop the predicted bounding boxes with an empirically determined margin of \SI{50}{px} to provide more context information. Next, we prompt a VLM with the full semantic description. If supported by the VLM, we further add a system prompt to explain the context of the task. If the vision language model confirms a proposal, we return the bounding box coordinates to the UAV and track the object until a human review is completed. The classification performance of the VLM appears to be significantly influenced by the choice of prompt. Since we test several VLMs of different sizes, we empirically tune the prompt for every model to optimize performance. We show an example for GPT4-mini:

\noindent {{\tt "role": "system", "content": "You are an intelligent AI assistant that helps a drone in a search and rescue mission. If in doubt, rather say yes."}\\
{\tt"role": "user", "content": "The drone needs to find an injured person. Would you recommend sending help to the person in the image? Start your justification with `Lets analyze the image'."}}



\subsection{Object Tracking Front-End}
\label{sec:method_frontend}
In SAR missions, targets are often in motion, such as lost children or individuals involved in maritime accidents. Therefore, a tracking system is essential to maintain an accurate, real-time position of any potential detection. 
A na\"ive approach is to stream every image from the UAV to the back-end and complete tracking on the back-end 
platform. 
Given adequate resources, this could either be the ground control station or a dedicated cloud server.
In many cases, data links are of limited bandwidth and potentially unstable. This leads to slow and unpredictable tracking frame rates, high delays in UAV course corrections and thus to the potential loss of an object of interest. 


We, therefore, require an onboard solution for short-term object tracking. This solution must support the OV nature of our pipeline. We further require the system to run with an adequate frame rate of at least \SI{10}{FPS} on common drone companion computers from the NVIDIA Jetson and Raspberry Pi series. To this end, we benchmark six compatible lightweight trackers on common drone hardware (Raspberry Pi 4 (2 Gb) and  NVIDIA Jetson Orin NX (16 Gb)). We evaluate each tracker on our two platforms and then select the best-performing tracker in terms of tracking accuracy. 

\subsection{Re-Initialization}
A front-end tracker can lose track in challenging conditions. We therefore require a re-initialization strategy. While some trackers provide self-evaluation through a tracking score, others do not offer this feature. In the case of score-enabled trackers, we tune a re-initialization threshold based on the VOT22 benchmark~\cite{Kristan2022a}.
We choose the threshold in such a way that performance in terms of the mean Intersection over Union (mIoU) is maximized (see Section \ref{sec:method_frontend} for further details).
For other trackers, we do not perform re-initialization.
In case of multiple detections during re-initialization, we pick the detection closest in position and dimension to the previously tracked box. We do so by minimizing the cost function $C_\text{box}$:

\begin{equation}
    C_\text{box}(p,i)=
    \sum_{d \in \{x, y\}} 
    \|c_{d,p}-c_{d,i}\|+
    \|s_{d,p}-s_{d,i}\| 
\end{equation}

Here, $p$ denotes the previously tracked box and $i$ is the incoming candidate for re-initialization. Further, $c$ denotes the center coordinate of the bounding box and $s$ its size. $d$~represents the $x$ and the $y$ coordinates 
in the pixel frame.

\section{DATASETS}
We evaluate our system based on two publicly available datasets, SARD~\cite{sambolek_search_2021} and VOT22~\cite{Kristan2022a}. In order to adapt the benchmarks for our use case, we apply the changes below and release the modified datasets as part of our work. 

\subsection{Search and Rescue Image Dataset for Person Detection (SARD)}
The ``Search and Rescue Image Dataset for Person Detection (SARD)''~\cite{sambolek_search_2021} consists of \num{6532}~instances of actors in different poses on \num{1981}~images. The images are taken from a UAV perspective in a natural setting. The dataset natively contains ground truth for the actors bounding boxes and poses (i.e., standing, lying, etc.). As part of this work, we annotate the color of each person's shirt. We approach this in a two-step manner. First, we prompt the GPT-4o API with a cropped image patch, containing the individual, for the color of the shirt. In the second step, we manually review the results for each individual. We then add ground truth for the recommendation for help. We first assume that all individuals requiring assistance are instances of the pose classes "laying\_down", "not\_defined", "null" and "seated". In many cases, classification based on the pose alone is not sufficient. We, therefore, classify all resulting \num{3255}~instances individually. We label individuals as "injured", if they are in a pose signalizing pain, injury, or unconsciousness. 

\subsection{Referring VOT22}
\label{sec:datasets}
The Visual Object Tracking (VOT) challenge~\cite{Kristan2022a} originally consists of \num{62}~sequences of a variety of objects and their ground truth positions. It assesses an algorithm's ability to track an object of interest across multiple frames. In the original benchmark, trackers are initialized with a segmentation mask. We manually create referring expressions for these objects and remove sequences where a unique description of the object of interest is not possible. This includes, for example, tracking a specific ant of a colony across frames. We call the augmented dataset \textit{Referring VOT22}.

\section{EXPERIMENTAL EVALUATION}
The main contribution of this paper is an OV object tracking method for SAR missions that runs online on commodity drone computers. We validate the approach in two steps. In Section \ref{sec:semantic_assessment}, we first evaluate the performance and open-vocabulary capabilities of our detection back-end on the SARD dataset and compare it with two state-of-the-art methods. In Section \ref{sec:end-to-end}, we then benchmark the tracking accuracy and speed of the full pipeline on the Referring VOT22 dataset using common UAV companion computers. 

Our approach is modular regarding the VLM in the back-end and the tracker method in the front-end. It also depends on a tracking re-initialization threshold. Throughout the experiments, we carry out sensitivity studies and a benchmarking of different component choices in varying settings. The corresponding results can assist meta-parameter selection for adapting our method to novel settings or hardware. 

\subsection{Hardware Set-up}
\label{section:hardware}
The proposed method requires a back-end server that runs the detection module and has a network connection to the UAV. In all experiments, we run the detection back-end on a workstation with an NVIDIA RTX A6000 GPU. For the tracking front-end, we deploy the two common UAV companion platforms Raspberry Pi 4 and NVIDIA Jetson Orin NX. 
The companion computer and the workstation are linked by a Gigabit Ethernet connection. We assume the drone to feature a 4G network with a bandwidth of at least \SI{5}{Mbps} in the field. We limit the connection to this speed in our tests.
\subsection{ OV Detection Evaluation}
\label{sec:semantic_assessment}
We first evaluate the detection back-end separately from the tracker. In particular, we benchmark the detection quality and speed of several back-end variants, each using a different VLM for final object classification, and compare against two baselines on the SARD dataset. 

First, we design four types of detection tasks suitable for SAR missions, each constituting a person's description with a different level of semantic complexity. By varying the person's cloth color and posture, we in total obtain \num{8}~referring expressions as detection objectives:

\begin{enumerate}
    \item Find any person.
    \item Find any person wearing a \{gray, green, blue\} shirt. 
    \item Find any person \{laying down, standing, sitting\}.
    \item Find any person, who might be injured and require assistance.
\end{enumerate}

Next, we configure our detection back-end, which consists of two parts. For the initial detection and bounding box generation, we choose Grounding DINO due to its good benchmark performance. For the fine-grained semantic classification, we evaluate GPT4-mini, LLaVA13b, LLaVA7b, and PaliGemma. All four configurations are prompted with the above detection objectives and applied on the SARD dataset. Since no tracking is performed in this evaluation part, the detector processes all images individually.

\begin{table}[t]
    \centering
    \caption{OV object detection evaluation. For each task and VLM, we report the Average Precision in the top part of the table.  We then calculate the mean among all experiments to obtain the mean Average Precision (mAP) metric for each VLM configuration. We further benchmark the per-frame $t_f$ and per-object $t_{obj}$ processing times.}
    \label{table:mapOurs}
    \tabOursAP
\end{table}

We evaluate the semantic detection results in mean Average Precision (mAP) and processing time. For the latter, we need to consider the back-end architecture.
The OV detector first generates bounding boxes for every instance of the superset class and
the VLM is then called individually for every detected bounding box. Thus, the runtime per frame scales linearly with the number of detected objects and depends on image content and detection prompt. We, therefore, report two temporal performance metrics. $t_{obj}$ describes the average runtime per detected instance and is approximately constant across scenarios. We also report the average processing time per frame $t_f$. This value is more common in literature and permits comparison with state-of-the-art approaches.

\paragraph*{VLM Benchmarking} Table~\ref{table:mapOurs} shows the object detection performance of our back-end in different configurations on different tasks.
PaliGemma as the smallest tested VLM performs well in semantically simple tasks, such as the detection of an unspecified person. However, it is not stable in more complex scenarios, such as the identification of a person by its pose, color of cloth or need for assistance.
Due to its smaller size, it exhibits the shortest runtime among all considered models. 

LLaVA7b overall performs better than PaliGemma, especially in the semantically more complex tasks. Interestingly, the mAP score diminishes for the simpler "person" identification scenario. Further analysis shows that the model is more conservative about its predictions, leading to the rejection of multiple true positives. This issue could possibly be addressed by prompt engineering in the future. The larger LLaVA13b marginally increases the performance by \SI{1}{\%} mAP over its smaller counterpart at the cost of higher runtime.

GPT4-mini delivers the best detection performance among all variants with \SI{49}{\%} mAP but also has the highest runtime.
Surprisingly, compared to LLaVA13b it scores lower in the "injury" class. On closer inspection, we found that the lower score was due to a higher number of false positives, resulting in a higher recall at the cost of lower mAP. 
Here, LLaVA13b scores \SI{73.28}{\%} recall, where GPT4-mini scores \SI{77.47}{\%}.
An explanation can be drawn from GPTs decision justifications. Contrary to other systems, GPT4-mini better understands the purpose of a SAR mission. While LLaVA mainly draws conclusions from the pose of the person, GPT4-mini tends to also recommend sending help in other scenarios. For example, when presented with an image of a  (not injured) child, GPT responds: 

\noindent{{\tt "The individual appears to be a young child standing alone, which may indicate they are lost or in need of assistance. Given the context of a SAR mission, it's prudent to send help to ensure the child's safety."}}

To conclude, we recommend GPT4-mini as a VLM when semantic understanding is critical and the back-end server has access to the internet. Otherwise, LLaVA 13b is a good alternative that can be run locally.

\paragraph*{Baseline Comparison} We compare our approach with GPT4-mini as VLM against GLEE \cite{wu2023GLEE} and Grounding DINO, two state-of-the-art OV object detection methods. For GLEE, we evaluate the versions with the largest backbone, GLEE plus and GLEE pro. Our method also uses Grounding DINO for initial object detection but with an additional verification step by a VLM. We evaluate the baselines on the same data and tasks as above.
As displayed in Table~\ref{table:mapBaseline}, none of the tested baselines can achieve a similar semantic understanding accuracy as our approach. Grounding DINO as our best-tested baseline performs comparably well on the task of detecting a person. 
It struggles, however, when prompted for semantically more complex scenarios. 
Further, its performance is not stable across several evaluations of the same semantic complexity. This becomes clear when comparing mAP scores on the "shirt\_gray" experiment with the performance on the "shirt\_blue" experiment. 
According to the authors, GLEE shows strong performance when tracking an object across several frames. But in their evaluation, the authors prompt the system with less semantically complex scenarios than used in our comparison. We believe this is a possible reason for its sub-average performance in our experiments. 

\begin{table}
    \centering
    \caption{OV object detection baseline comparison. Here, \algName~ is our method with GPT4-mini as VLM. We calculate the Average Precision score (AP) for each experiment. We then calculate the mean among all experiments to obtain the mean Average Precision (mAP) metric for each baseline. 
    \algName~performs best with an mAP score of~\SI{49.39}{\%}.
    }
    \label{table:mapBaseline}
    \tabBaselineAP
\end{table}

\subsection{Full-Pipeline Evaluation}
\label{sec:end-to-end}
We next analyze the performance of our complete pipeline on two UAV companion computers and hardware set-up as described in Section \ref{section:hardware}. As the back-end, we deploy the configuration using LLaVA13b as VLM. Our main goal is to demonstrate that our OV tracking approach can run in real-time on this limited hardware. In the process, we also test several tracking front-end choices on both companion platforms. We conduct the evaluation on the Referring VOT22 benchmark (see Section~\ref{sec:datasets}) to obtain use case agnostic numbers due to a wider range of object classes and more complex object trajectories compared to SARD. 

Since SAR scenarios usually lead to a single-object tracking task after successful detection, we evaluate our results using the mean Intersection over Union (mIoU) metric. It is computed by summing the IoU for all detections and dividing it by the total number of ground truth objects. 
We further report the average back-end response time ($t_b$), the tracking frame rate on the edge platform ($\text{FPS}_\text{Edge}$) as well as the overall average frame rate. 




\begin{figure*}
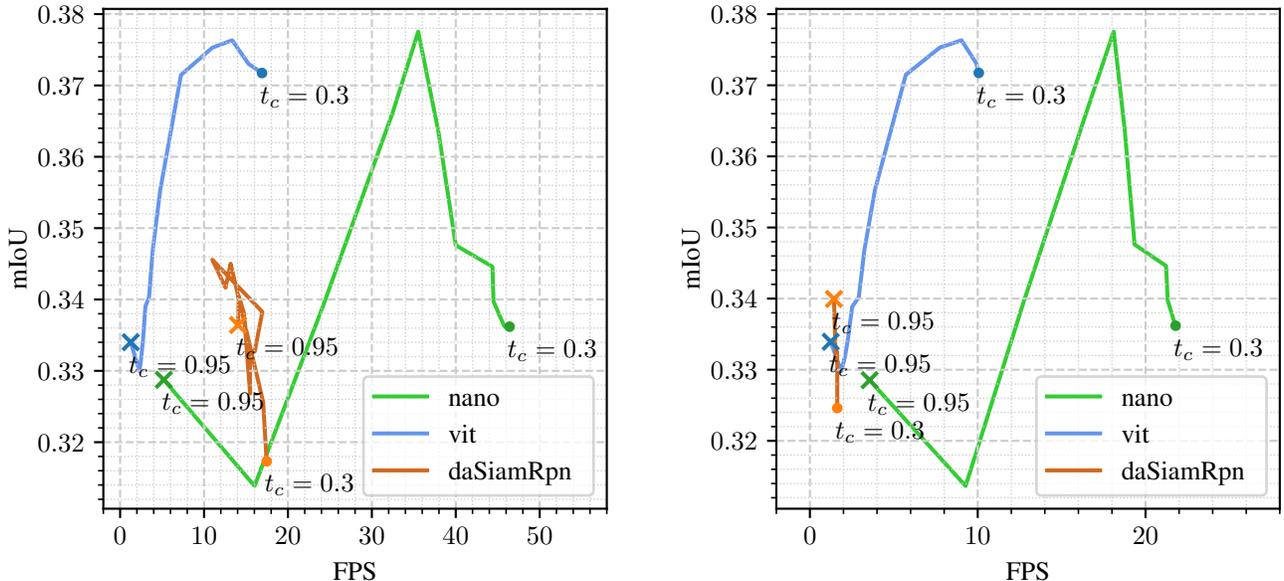

\begin{subfigure}{0.45\textwidth}
    \input{jetson.pgf}
    \label{fig:orin}
\end{subfigure}
\hspace{7mm}
\begin{subfigure}{0.45\textwidth}
    \input{pi.pgf}
    \label{fig:RaspPi}
\end{subfigure}
\caption{Sensitivity study for performance in terms of mean IoU (mIoU) and FPS under variation of the confidence threshold~$t_c$ for re-initialization. We analyze threshold-enabled trackers on a \textbf{Jetson Orin NX (left)} and a \textbf{Raspberry Pi 4 (right)} platform.}
\label{fig:diagrams}
\end{figure*}

We test six lightweight object trackers (CSRT, DaSiamRPN, GOTURN, mil, nano, and vit) on both companion platforms to find the best configuration. In the case of threshold-enabled trackers, we run a sensitivity study to choose the re-initialization threshold $t_c$ so that FPS and mIoU are maximized. We, therefore, vary $t_c$ between \num{0.3} and \num{0.95} in steps of \num{0.05} (see Figure \ref{fig:diagrams}). We conduct this study on the Referring VOT22 benchmark and report the tracking performance with best threshold value in Table~\ref{tab:endToEnd}.

We start analyzing the results of the Jetson Orin NX platform.
Among all tested alternatives, the configuration including the nano tracker performs best both on mIoU and overall FPS metric. The achieved edge tracking frame rate reaches \SI{66.067}{FPS}. 
Based on our evaluation, we recommend the threshold $t_{c,\text{opt}}$ for the nano tracker to be set to~\num{0.7}. Vit shows a similar tracking performance in terms of mIoU, scoring only slightly lower than nano. 
The $\text{FPS}_\text{Edge}$, however, lies \SI{43.9}{\%} below the one achieved by nano at a similar mIoU performance. But vit shows a more stable relation between mIoU and $t_c$. This could indicate a better calibration of its confidence score. 
Since $\text{FPS}_\text{Edge}$ for vit still exceeds the requirements for real-time processing, this configuration could also be applicable in real-time scenarios.
All trackers show similar mIoU scores across both platforms. The Jetson Orin NX platform enables higher frame rates due to its CUDA acceleration. Still, even on the relatively inexpensive Raspberry Pi 4 it is possible to run our method in real-time with the nano tracker, which delivers an $\text{FPS}_\text{Edge}$ of~\SI{29.487}{FPS}.
 

\begin{table}
    \centering
    \caption{Results with fine-tuned threshold values from the sensitivity studies on the Jetson Orin NX and Raspberry Pi~4 platforms.}
    \label{table:front-end}
\begin{tabular}{llllll}
\toprule
 & Tracker & $t_{c,\text{opt}}$ & mIoU $\uparrow$ & FPS $\uparrow$ & $\text{FPS}_\text{Edge} \uparrow$ \\
\midrule
 \parbox[t]{2mm}{\multirow{6}{*}{\rotatebox[origin=c]{90}{Jetson Orin NX}}}
& nano & 0.7 & \textbf{0.378} & \textbf{35.526} & \textbf{66.067} \\
& vit & 0.4 & 0.376 & 13.355 & 37.099 \\
& DaSiamRPN & 0.9 & 0.346 & 10.988 & 31.263 \\
& csrt & --- & 0.229 & 17.586 & 26.864 \\
& mil & --- & 0.153 & 2.69 & 3.019 \\
& GOTURN & --- & 0.091 & 16.875 & 22.602 \\
\midrule
\parbox[t]{2mm}{\multirow{6}{*}{\rotatebox[origin=c]{90}{Raspberry Pi 4}}}
& nano & 0.7 & \textbf{0.378} & \textbf{18.101} & \textbf{29.487} \\
& vit & 0.4 & 0.376 & 9.034 & 20.442 \\
& DaSiamRPN & 0.9 & 0.34 & 1.434 & 2.122 \\
& csrt & --- & 0.229 & 7.276 & 9.704 \\
& mil & --- & 0.153 & 6.034 & 7.202 \\
& GOTURN & --- & 0.091 & 2.722 & 3.006 \\
\bottomrule
\end{tabular}
\label{tab:endToEnd}
\end{table}

\section{CONCLUSION AND FUTURE WORK}
In this paper, we introduce \algName, a scalable OV object tracking approach with advanced semantic understanding. Our approach combines powerful foundation models in a cloud back-end with a real-time capable tracker on the UAVs companion computer. We extensively evaluate our framework on two common UAV companion computers, four different VLMs, and six object trackers to find the best-suited configuration for each hardware platform. As we show in our evaluation, our framework significantly outperforms other approaches in terms of semantic understanding. This, however, comes at the cost of increased runtime in the back-end. Future work could tackle this problem by fine-tuning the OV object detector or combining the results of fast and slow VLMs to reduce the latency time of the UAV. Further, multi-object tracking could be supported to permit use cases beyond SAR missions with more than one object of interest.






\bibliographystyle{IEEEtran}
\bibliography{sources.bib}




\end{document}

%% file: python_results.tex
\newcommand{\tabOursAP}{\begin{tabular}{lllll}
\toprule
\tikz{\node[below left, inner sep=1pt] (def) {Task};%
      \node[above right,inner sep=1pt] (abc) {VLM};%
      \draw (def.north west|-abc.north west) -- (def.south east-|abc.south east);}      & GPT4-mini   & LLaVA13b   & LLaVA7b   & PaliGemma   \\
\midrule
 person       & 75.23\%            & \textbf{82.95\%}           & 74.96\%          & 82.88\%           \\
 shirt\_gray   & \textbf{51.61\%}            & 44.96\%           & 38.38\%          & 40.35\%           \\
 shirt\_green  & \textbf{53.13\%}            & 40.62\%           & 37.53\%          & 3.7\%             \\
 shirt\_blue   & \textbf{54.76\%}            & 34.04\%           & 35.72\%          & 7.85\%            \\
 pose\_laying  & \textbf{42.77\%}            & 28.29\%           & 40.66\%          & 13.77\%           \\
 pose\_standing & \textbf{49.4\%}             & 45.9\%            & 39.9\%           & 39.97\%           \\
 pose\_sitting & \textbf{30.08\%}            & 21.18\%           & 25.57\%          & 9.22\%            \\
 injury       & 38.1\%             & \textbf{40.38}\%           & 37.4\%           & 4.92\%            \\
 \midrule
   mAP                 & \textbf{49.39\%}            & 42.29\%           & 41.26\%          & 25.33\%           \\
 \midrule
  $t_\text{f}$  & 6.955s            & 4.915s           & 3.856s          & \textbf{0.866s}           \\
 $t_\text{obj}$ & 2.239s            & 1.478s           & 1.16s           & \textbf{0.26s}            \\
\bottomrule
\end{tabular}}
\newcommand{\tabBaselineAP}{\begin{tabular}{lllll}
\toprule
 \tikz{\node[below left, inner sep=1pt] (def) {Task};%
      \node[above right,inner sep=1pt] (abc) {Method};%
      \draw (def.north west|-abc.north west) -- (def.south east-|abc.south east);}            &\algName                & gDINO             & GLEE plus   & GLEE pro   \\
\midrule
 person         & \textbf{75.23\%}       & 72.6\%            & 34.46\%           & 30.99\%          \\
 shirt\_gray    & \textbf{51.61\%}       & 24.79\%           & 29.61\%           & 21.09\%          \\
 shirt\_green   & 53.13\%                & 31.74\%           & \textbf{54.67\%}  & 35.63\%          \\
 shirt\_blue    & \textbf{54.76\%}       & 16.48\%           & 17.73\%           & 18.05\%          \\
 pose\_laying   & \textbf{42.77\%}       & 17.45\%           & 42.4\%            & 37.06\%          \\
 pose\_standing & \textbf{49.4\%}        & 27.33\%           & 15.78\%           & 19.43\%          \\
 pose\_sitting  & \textbf{30.08\%}       & 14.11\%           & 4.17\%            & 5.61\%           \\
 injury         & 38.1                   & 37.05\%           & 27.3\%            & \textbf{38.36\%}          \\
\midrule
 mAP            & \textbf{49.39\%}       & 30.19\%           & 28.27\%          & 25.78\%          \\
\midrule
 $t_\text{f}$   & 6.955s             & \textbf{0.284s}   & 0.319s           & 1.156s          \\
 $t_\text{obj}$ & 2.239s             & \textbf{0.103s}   & 0.265s           & 0.666s          \\
\bottomrule
\end{tabular}}

\newcommand{\tabFrontend}{
\begin{tabular}{lllll}
\toprule
 & Run & IoU & FPS & free_fps \\
\midrule
nano & 21 & 0.380000 & 35.530000 & 35.530000 \\
vit & 28 & 0.380000 & 13.360000 & 13.360000 \\
daSiamRpn & 12 & 0.350000 & 10.990000 & 10.990000 \\
csrt & 39 & 0.230000 & 17.590000 & 17.590000 \\
mil & 41 & 0.150000 & 2.690000 & 2.690000 \\
goturn & 40 & 0.090000 & 16.880000 & 16.880000 \\
\bottomrule
\end{tabular}
\end{tabular}}